\documentclass[lettersize,journal]{IEEEtran}

\usepackage{subfigure}
\usepackage{amsmath,amsfonts}
\usepackage{algorithmic}
\usepackage{algorithm}
\usepackage{array}
\usepackage[caption=false,font=normalsize,labelfont=sf,textfont=sf]{subfig}
\usepackage{textcomp}
\usepackage{stfloats}
\usepackage{url}
\usepackage{verbatim}
\usepackage{graphicx}
\usepackage{cite}

\usepackage{algorithm} 
\usepackage{algorithmic} 
\usepackage{bm}
\usepackage{multirow}
\usepackage{booktabs}
\usepackage{bbding}

\usepackage[switch]{lineno}

\begin{document}

\title{Branches Mutual Promotion for End-to-End Weakly Supervised Semantic Segmentation}

\author{Lei~Zhu, Hangzhou He, Xinliang Zhang, Qian~Chen, Shuang~Zeng, Qiushi~Ren, Yanye~Lu*}

\markboth{Journal of \LaTeX\ Class Files,~Vol.~14, No.~8, August~2021}%
{Shell \MakeLowercase{\textit{et al.}}: A Sample Article Using IEEEtran.cls for IEEE Journals}

\maketitle

\begin{abstract}
End-to-end weakly supervised semantic segmentation aims at optimizing a segmentation model in a single-stage training process based on only image annotations. Existing methods adopt an online-trained classification branch to provide pseudo annotations for supervising the segmentation branch. However, this strategy makes the classification branch dominate the whole concurrent training process, hindering these two branches from assisting each other. In our work, we treat these two branches equally by viewing them as diverse ways to generate the segmentation map, and add interactions on both their supervision and operation to achieve mutual promotion. For this purpose, a bidirectional supervision mechanism is elaborated to force the consistency between the outputs of these two branches. Thus, the segmentation branch can also give feedback to the classification branch to enhance the quality of localization seeds. Moreover, our method also designs interaction operations between these two branches to exchange their knowledge to assist each other. Experiments indicate our work outperforms existing end-to-end weakly supervised segmentation methods. 
\end{abstract}


\section{Introduction}
\label{sec:1}
\begin{figure}
\centering
\includegraphics[width=0.50\textwidth]{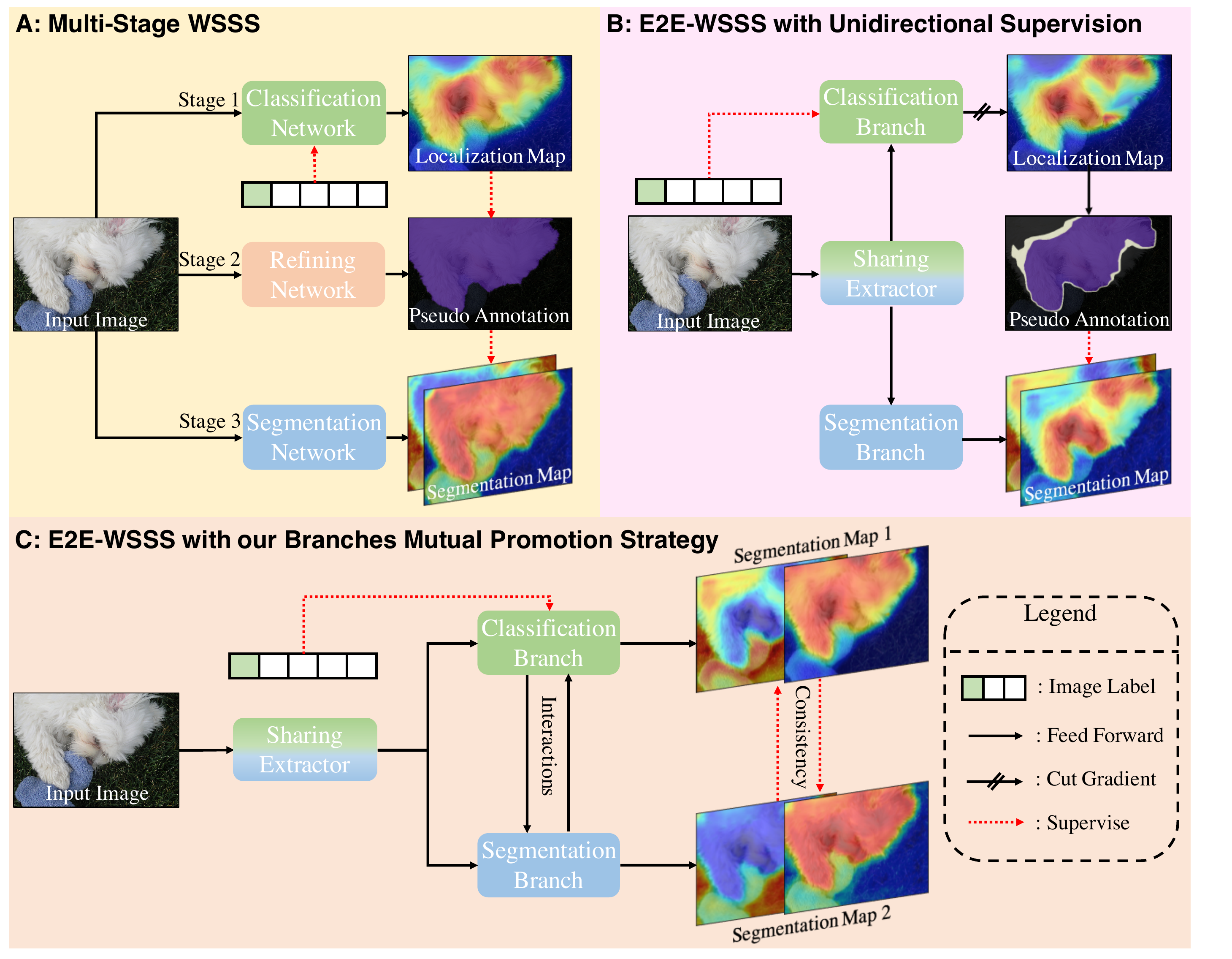}
\caption{Comparison of WSSS strategies:  \textsf{A}. Multi-stage WSSS contains multiple training stages. \textsf{B}. Existing E2E-WSSS unidirectionally supervises the segmentation branch with pseudo annotations online provided by the classification branch.  \textsf{C}. Our proposed E2E-WSSS strategy interacts both the supervision and operation between these two branches to achieve mutual promotion.}
\label{fig:intro_1}
\end{figure}

\IEEEPARstart{S}{emantic} segmentation is a primary vision task, aiming to annotate pixels in an image as target objects or backgrounds. However, training a segmentation model in a fully-supervised manner requires annotating all pixels in training images, costing extensive human resources. To solve this problem, weakly supervised semantic segmentation (WSSS) appears and attracts extensive attention, which adopts only image-level annotation for the training process. However, as shown in Fig.~\ref{fig:intro_1} \textsf{A}, WSSS methods usually require multiple training stages, \textit{e.g.}, tuning a classification network with image annotations to produce localization seeds~\cite{CAM, ADL, SEAM}, deriving pseudo annotations after refining the seeds~\cite{AFF, IRN, CIAN}, and finally training the segmentation network with the pseudo annotations~\cite{DEEPLAB, Segformer}.

Recently, some end-to-end weakly supervised semantic segmentation (E2E-WSSS) methods arose to simplify the heavy multi-stage training process into a single stage~\cite{RRM, RRMv2, AFA, AALR}. As shown in Fig.~\ref{fig:intro_1} \textsf{B}, these methods train a two-branch network in only a single stage, where the classification branch supervised by image-level annotation can online provide pseudo annotations for the segmentation branch. Compared with multi-stage WSSS, the concurrently-trained classification branch cannot stably provide seed to derive accurate pseudo annotations for supervising the segmentation branch. So, existing E2E-WSSS methods focus on improving the classification branch to provide better supervision by refining the localization seed with online spatial propagation~\cite{CRF,AFA, AALR} or determining reliable regions on the pseudo annotations~\cite{RRM, RRMv2}. 

\begin{figure}
\centering
\includegraphics[width=0.49\textwidth]{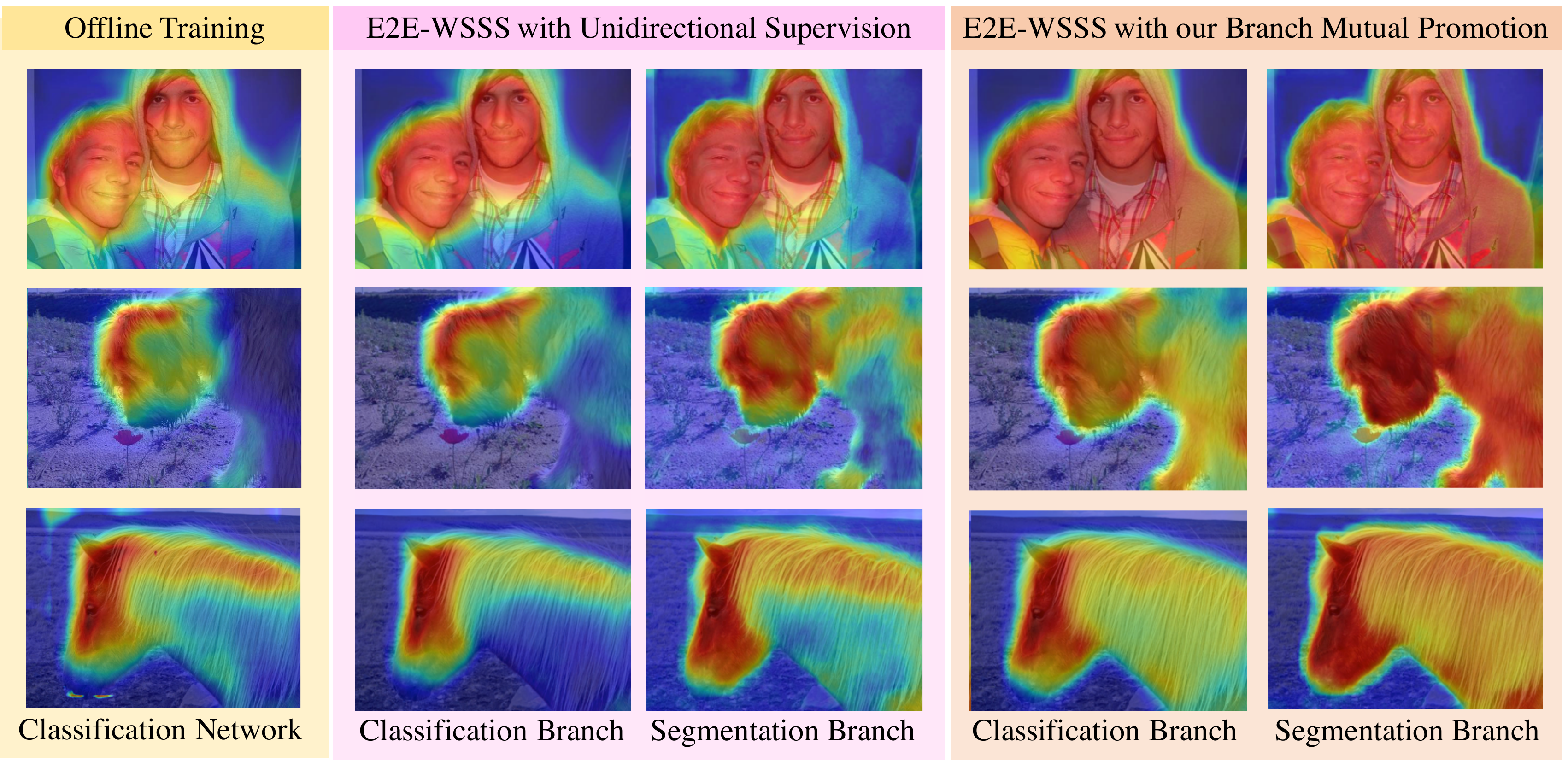}
\caption{Localization maps generated by different strategies. The classification branch trained with unidirectional supervision (noted by purple) cannot profit by the segmentation branch, having similar activations as the offline-trained network (noted by yellow). Our method (noted by orange) enable segmentation branch give feedback to the classification branch for mutual promotion.}
\label{fig:intro_2}
\end{figure}

In our work, we argue that current E2E-WSSS methods may fall into a trap, following the multi-stage WSSS to unidirectionally supervise the segmentation branch based on the prediction of the classification branch, without considering the feedback of the segmentation branch. In this way, the classification branch will dominate the whole training process, even if it may perform worse than the segmentation branch, as visualized in Fig.~\ref{fig:intro_2}. Thus, the classification branch will converge to a similar optimum as the offline trained classification network but cannot stably provide pseudo annotations for the segmentation branch, which causes the large performance gap between current E2E-WSSS and multi-stage WSSS methods.

Actually, in the E2E-WSSS setting, these two branches are basically at equal status because they are concurrently optimized during training. From another perspective, the segmentation branch can also assist the concurrently-trained classification branch in generating better localization seeds, which is a crucial trait of E2E-WSSS and yet to be explored by existing methods. Based on this perspective, our work treats these two branches equally by viewing them as diverse ways to achieve the same goal, generating the segmentation map of input images. Thus, as shown in Fig.~\ref{fig:intro_1} \textsf{C}, interactions are elaborated on both their supervision and operations to promote them mutually. For this purpose, our approach chooses to regularize the consistency between these two branches with cross-supervision to consider their identical target for training. This interactional supervision enables the segmentation branch to give feedback to the classification branch to improve its performance and vice versa. Moreover, based on the specificities of these two branches, interactional operations are designed to exchange their knowledge, such as providing object and background priors. In this way, these two branches can work cooperatively to better segment input images with only the supervision of image-level annotations.

In a nutshell, our contributions are fourfold:
\begin{itemize}
\item {Our work rethinks the relation between the two branches of E2E-WSSS, and suggest to interactively operate them for mutual promotion}
\item {An interactional supervision mechanism is proposed to enable mutual feedbacks between two branches with cross-task pseudo supervisions.}
\item {Branch interaction mechanisms are presented to interact the operations between these two branches to exchange their knowledge.}
\item{Experiments indicate that our method outperforms E2E-WSSS methods and is even comparable with multi-stage methods on VOC2012 and MS-COCO dataset.}
\end{itemize}

\section{Related Work}
\noindent \textbf{Multi-stage WSSS methods:} Multi-stage WSSS methods adopt multiple stages to train the segmentation model with image-level annotation. The majority of them pay attention to improving the quality of localization seeds by tuning the classification network with different approaches, \textit{e.g.}, using erasing-based mechanisms to erase discriminative parts during training~\cite{ADL, CSE, ECS}, designing regularizations to force the network to catch undiscriminating cues~\cite{SEAM, SIPE, CPN}, and engaging auxiliary tasks for assistance~\cite{ICD, AdvCAM, ReCAM, SCCAM, MMCST,CLIMS}. In addition, some works also focus on further refining the localization seed to produce high-quality pseudo annotations with refinement networks, trained based on confident seed regions to predict pixel affinity matrixes~\cite{AFF, CIAN} or boundary maps~\cite{IRN, BES}. Moreover, strategies for better training the segmentation network with noisy pseudo annotations are also explored for the multi-stage WSSS method~\cite{ADELE}.  Profited by the stability of the off-line generated pseudo annotations, multi-stage WSSS methods usually have satisfactory performance. However, this type of method usually contains more than three training stages, which also disable the interactions between their object localization and segmentation structures.


\noindent \textbf{End-to-end WSSS methods:} End-to-end WSSS methods learn the segmentation model with image-level annotations in only one stage. Some earlier E2E-WSSS works adopt a single-branch network to produce both segmentation maps and image-level classification scores. Typically, Pedro \textit{et al.}~\cite{MIL} adopt multi-instance learning (MIL) to train the classification network for creating pixel-level segmentation maps with additional local priors such as superpixel~\cite{FH}. Roy \textit{et al.}~\cite{CRNN} introduce a CRF-CNN to regularize and refine the segmentation map outputted by the single-branch network trained with MIL. Pinheiro \textit{et al.}~\cite{1Stage} elaborate a normalized global weighted pooling and pixel-adaptive mask refinement to self-train the segmentation network with image annotations. However, the inconsistent distribution between image-level and pixel-level features~\cite{DAWSOL} limits the performance of the single-branch E2E-WSSS methods. Thus, current E2E-WSSS methods usually follow the seed-expand-constrain principle~\cite{SEC} to concurrently train a classification network and a segmentation network, respectively for generating the localization seed and segmentation map. Zhang \textit{et al.}~\cite{RRM, RRMv2} share the extractor of these two networks to form a two-branches structure for training and select reliable locations in localization seeds as the pseudo annotation with the help of the conditional random fields (CRF)~\cite{CRF}. Zhang \textit{et al.}~\cite{AALR} design an adaptive affinity loss based on the connectivity of pseudo annotation to regularize semantic propagation. Ru \textit{et al.}~\cite{AFA} propose the affinity form attention (AFA) structure, which directly learns affinity from the self-attentions outputted by the transformer backbone. However, these methods all unidirectionally train the segmentation branch with the pseudo annotation provided by the classification branch, making the performance of the segmentation branch choked by the concurrently-trained classification branch. Different from them, our method treats these two branches equally and adds interaction to make them achieve mutual promotion.


\begin{figure*}
\centering
\includegraphics[width=0.98\textwidth]{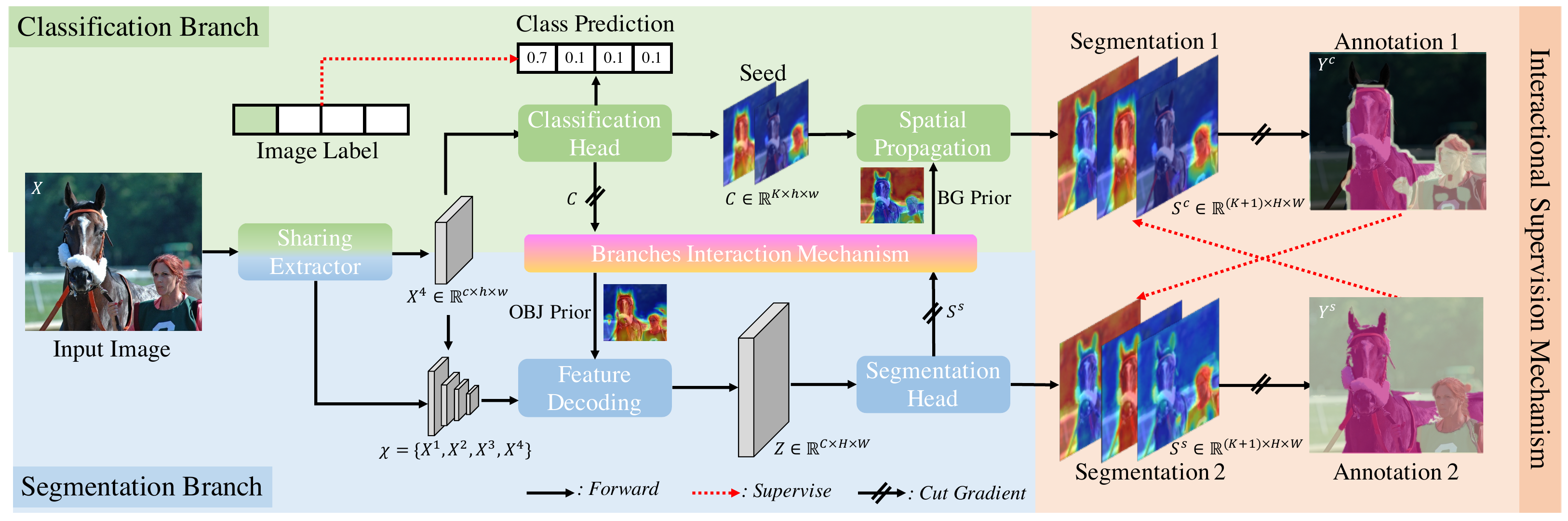}
\caption{Overview of our proposed methods. An interactional supervision mechanism (noted by orange) is proposed to enforce the consistency between the classification branch (noted by green) and the localization branch (noted by blue) with cross-supervision. Moreover, the branches interaction mechanism also adopts to exchange the object (OBJ) prior and background (BG) prior between these two branches.}
\label{fig:overview}
\end{figure*}

\section{Method}
This section first defines the problem of training a segmentation model with image-level annotation in an end-to-end manner. Then, we explore the relation of the two branches in E2E-WSSS and conclude that they act as different means for generating segmentation maps. Next, mechanisms designed based on the relations between these two branches to add interactions for their supervision and operation are respectively proposed in Sec.~\ref{sec:33} and Sec.~\ref{sec:34}. Finally, we introduce the network structure and workflow for training with our interaction mechanisms.  

\subsection{Problem Definition}
Given an input image $\bm{X} \in \mathbb{R}^{3 \times H \times W}$, semantic segmentation aims at classifying pixels in the image with a segmentation network, where $H \times W$ is the resolution of the image. To achieve this purpose, the segmentation network needs a feature extractor $e(\cdot)$ to extract multi-level features:
\begin{equation}
\mathcal{X} = \{ \bm{X}^{1}, \bm{X}^{2}, \bm{X}^{3}, \bm{X}^{4} \} = e(\bm{X})~~,
\end{equation}
\noindent where $\bm{X}^{i}$ represents the $i^{th}$-level feature, and we call $\bm{X}^{4}$ the high-level feature. Then, a feature decoder $d(\cdot)$ is used to fuse $\mathcal{X}$ into $\bm{Z} \in \mathbb{R}^{C \times H \times W}$ to consider multi-level cues. This fused feature $\bm{Z}$ is finally classified by a segmentation head $s(\cdot)$ to generate the output segmentation map $\bm{S}^{s} \in \mathbb{R}^{ (K+1) \times H \times W}$, representing the probability that pixels belong to the background class and the $K$ target objects.

To end-to-end train the segmentation network with weak supervision, a classification network, sharing the feature extractor $e(\cdot)$, is required to produce the pseudo annotation of the segmentation network. It utilizes a classification head $c(\cdot)$ and a global max-pooling $g(\cdot)$ to output an image-level prediction $\bm{c} = c( g(\bm{X}^{4}) ) \in \mathbb{R}^{K \times 1}$, which is fully supervised by the image-level annotation $\bm{y} \in \mathbb{R}^{K\times1}$. This classification head is also projected onto the high-level feature $\bm{X}^{4}$ to generate a coarse localization seed $\bm{C} = c(\bm{X}^{4}) \in \mathbb{R}^{K \times h \times w}$, which should be further upsampled into image resolution $H \times W$ and diffused with the off-the-shelf propagation $p(\cdot)$ to consider low-level cues. Finally, an additional background score $\bm{B} \in \mathbb{R}^{1 \times H \times W}$ is also utilized to filter the seed to produce the pseudo annotation $\bm{Y}^{c}$ as the supervision of the segmentation network. For clarity, in the rest of paper, we name the detached parts of these two networks as the segmentation branch and the classification branch.

\subsection{Rethinking the Branches in E2E-WSSS}
The workflow of the classification branch can also be reformulated by wrapping the localization map with the background score to form another segmentation map $\bm{S}^{c} \in \mathbb{R}^{ (K+1) \times H \times W}$ before determining the pseudo annotation $\bm{Y}^{c}$. From this perspective, the classification branch and segmentation branch can be summarized as two diverse ways to produce the segmentation map of the input image:
\begin{align}
\begin{split}
\bm{X} \rightarrow e(\cdot) \rightarrow d(\cdot) \rightarrow s(\cdot) \rightarrow \bm{S}^{s}& \\
~~~~~~~~~~~~~~~~\searrow c(\cdot) \rightarrow p(\cdot) \rightarrow \bm{S}^{c}& \leftarrow \bm{B} 
\end{split} ~~,
\label{eq:workflow}
\end{align}
\noindent Specifically, the segmentation branch contains a unique feature decoding step $d(\cdot)$, which can fuse the multi-level features to consider different-level cues before being classified by the segmentation head $s(\cdot)$. Benefiting from this, the segmentation map $\bm{S}^{s}$ generated by the segmentation branch will have smooth activation and contain more detailed structures. However, due to the lack of precise supervision, this feature decoding process also introduces irrelevant fusion, limiting its ability to distinguish different classes of objects.

Different from the segmentation branch, the classification branch directly classifies the high-level feature $\bm{X}^{4}$ by the object classification head $c(\cdot)$ to produce coarse localization seeds, containing only the object score with feature-level resolution. These coarse seeds are then spatially diffused with $p(\cdot)$ to consider low-level cues, such as image density~\cite{1Stage, CRF} and spatial position~\cite{AFA}. Profited by the utilization of high-level features and the additional image-level supervision, the segmentation map $\bm{S}^{c}$ produced by the classification branch usually better discerns the location of different objects. However, $\bm{S}^{c}$ also requires additional background score $\bm{B}$ and suffers from unsmooth activation, because the object classification head $c(\cdot)$ cannot identify the background class and lacks multi-level consideration.

Concluded from the above analysis, these two branches have the same goal with relatively complementary specificities caused by their different workflows. So, unlike other E2E-WSSS methods that operate the two branches separately, our approach concerns their relationships to achieve mutual promotion. Based on their relations, interactions are added to both their supervision and operations, as shown in Fig.~\ref{fig:overview}. In this way, these two branches can work corporately rather than independently to improve their performance.


Firstly, to meet the above purpose, an interactional supervision mechanism is designed to consider the similarity between these two branches, \textit{i.e.}, their same target of producing high-quality segmentation maps. As generalized in Fig.~\ref{fig:overview} (noted by orange), this mechanism enhances the consistency between these two branches with cross-pseudo supervision~\cite{CPS} to enable their interaction, rather than unidirectionally forcing the segmentation branch to follow the classification branch. In this way, the segmentation map $\bm{S}^{s}$ can also provide feedback to the classification branch to assist $\bm{S}^{c}$ in generating smooth and detailed object activations.

Moreover, a branches interaction mechanism is also elaborated to consider the specificities caused by the different workflows of these two branches. This mechanism adds interactive operations between these two branches to exchange their knowledge. As shown in Fig.~\ref{fig:overview}, object and background priors can be exchanged by our branch interaction mechanism to reduce the irrelevant feature fusion for the segmentation branch and offset the background unawareness for the classification branch, respectively.

\subsection{Interactional Supervision Mechanism}
\label{sec:33}
The proposed interactional supervision mechanism assumes that the outputs of the classification and segmentation branches should be consistent in theory due to their same target. For this purpose, the consistency between these two branches is enforced by our mechanism to enable their interaction during the end-to-end training process, rather than just using $\bm{Y}^{c}$ to supervise the segmentation branch unidirectionally as existing methods~\cite{AFA,RRM,RRMv2,AALR,BES}. 

As shown in Fig.~\ref{fig:overview} (orange part), our mechanism views the classification and the segmentation branch as two different models for producing segmentation maps, whose consistencies can be regulated by cross-pseudo supervision~\cite{CPS}. Thus, two pseudo annotations are generated by the maximum predictions of these two parallel branches: 
\begin{equation}
\bm{Y}^{c}_{i,j} = \arg\max (\bm{S}^{c}_{:, i, j})~~,~~
\bm{Y}^{s}_{i,j} = \arg\max (\bm{S}^{s}_{:, i, j})~~.
\end{equation}

\noindent These pseudo annotations are used to supervise the segmentation map of the other branch to regulate the consistency:
\begin{equation}
\mathcal{L}_{is} = \mathcal{L}_{c2s}(\bm{S}^{s}, \bm{Y}^{c}) + \mathcal{L}_{s2c}(\bm{S}^{c}, \bm{Y}^{s})~~,
\label{eq:bd_loss}
\end{equation}
\noindent where $\mathcal{L}_{is}$ is the objective of our interactional supervision.

\begin{figure*}
\centering
\includegraphics[width=0.99\textwidth]{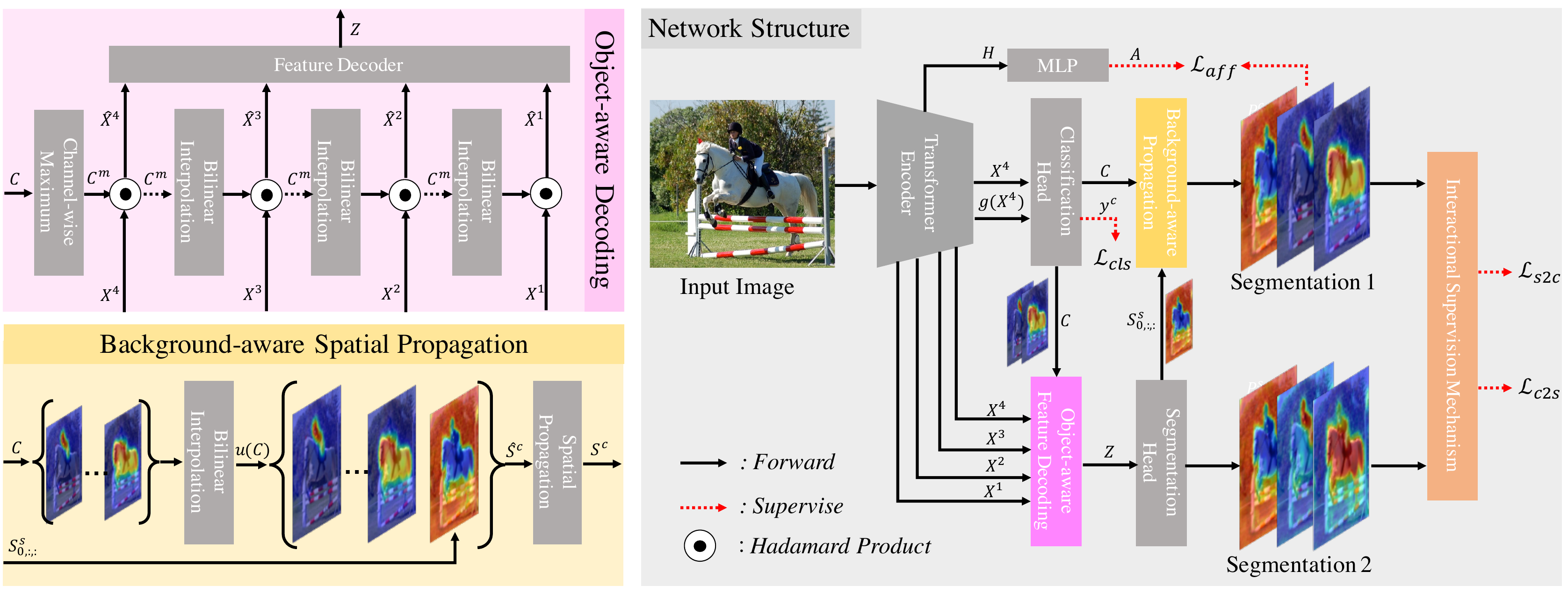}
\caption{Workflows of the object-aware decoding operation (noted by purple), the background-aware spatial propagation operation (noted by yellow), and the total network for training with our method (noted by gray). }
\label{fig:structure}
\end{figure*}

In detail, the first term $\mathcal{L}_{c2s}$ uses the classification branch to generate the pseudo annotation for supervising the segmentation branch with a mask matrix that filters the unconfident spatial location:
\begin{align}
\begin{split}
\bm{M}^{c}_{i, j} = \left \{
\begin{array}{ll}
1 , & \max(\bm{S}^{c}_{:, i, j}) > \sigma^{c} \\
0 , & \rm{otherwise}
\end{array}
\right.
\end{split}~~,
\label{eq:relation}
\end{align}
\noindent where $\sigma^{c}$ is a threshold. Then, the segmentation loss between the pseudo annotation $\bm{Y}^{c}$ and the segmentation map $\bm{S}^{s}$ on the reliable regions marked by $\bm{M}^{c}$ is computed:
\begin{equation}
\mathcal{L}_{c2s} = \frac{1}{|\bm{M}^{c}|} \sum^{H,W}_{i, j=1} \bm{M}^{c}_{i,j} * l_{seg}(\bm{S}^{s}_{:, i, j}, \bm{Y}^{c}_{i, j})~~,
\end{equation}
\noindent where $l_{seg}(\cdot)$ can be implemented as the multi-label soft margin loss or the cross-entropy loss. Like the other methods~\cite{AFA,RRM,RRMv2, AALR}, this term forces the segmentation branch to catch the object locations discerned by the classification branch to provide reliable segmentation maps $\bm{S}^{s}$.

In addition, the other term $\mathcal{L}_{s2c}$ is adopted as the feedback of the segmentation branch, which uses the pseudo annotation $\bm{Y}^{s}$ to supervise the segmentation map $\bm{S}^{c}$ outputted by the classification branch. Similarly, a mask matrix is also generated for filtering:
\begin{align}
\begin{split}
\bm{M}^{s}_{i, j} = \left \{
\begin{array}{ll}
1 , & \bm{Y}^{s}_{i, j} \neq 0 ~, ~ \max(\bm{S}^{s}_{:, i, j}) > \sigma^{s} \\
0 , & \rm{otherwise}
\end{array}
\right.
\end{split}~~,
\label{eq:relation}
\end{align}
\noindent where $\sigma^{s}$ is also a confidence threshold. Different from $\bm{M}^{c}$, mask $\bm{M}^{s}$ also filters the background regions because the background class in $\bm{S}^{c}$ is provided by the off-the-shelf background prior as indicated in Eq.~\ref{eq:workflow}. Based on $\bm{M}^{s}$, the segmentation loss between $\bm{S}^{c}$ and $\bm{Y}^{s}$ on marked regions is adopted as the feedback loss term $\mathcal{L}_{s2c}$:
\begin{equation}
\mathcal{L}_{s2c} = \frac{1}{|\bm{M}^{s}|} \sum^{H,W}_{i, j=1} \bm{M}^{s}_{i,j} * l_{seg}(\bm{S}^{c}_{:, i, j}, \bm{Y}^{s}_{i, j})~~.
\end{equation}
\noindent Benefiting from $\mathcal{L}_{s2c}$, the segmentation branch can also take effect to the classification branch, helping $\bm{S}^{c}$ generate a smooth activation to cover more object parts. The quality improvement of $\bm{S}^{c}$ will then transmit to $\bm{Y}^{c}$ to provide more accurate pseudo annotations for optimizing $\bm{S}^{s}$ with $\mathcal{L}_{c2s}$, making these two branches achieve mutual promotion. From another point, this additional term also enables the classification head, supervisedly trained based on image-level features, perceive the distribution of pixel-level features to enhance the quality of localization seeds~\cite{DAWSOL}. 

\subsection{Branches Interaction Mechanism}
\label{sec:34}
Besides considering the similarities of the classification and segmentation branches, our method also utilizes their specificities to elaborate interaction operations that benefited them from the knowledge provided by the other.


\noindent \textbf{Object-aware Feature Decoding:} Firstly, to compensate for the object unawareness in the feature decoding step, our method designs an object-aware feature decoding (OFD) operation that utilizes the localization seed $\bm{C}$ of the classification branch to provide object priors for the segmentation branch. As shown in Fig.~\ref{fig:structure} (noted by purple), OFD first generates the maximum object score based on $\bm{C}$ to represent the probability that a spatial position contains objects:
\begin{equation}
\bm{C}^{m}_{i, j} = \max(\bm{C}_{0, i, j}, \bm{C}_{1, i, j}, \cdots, \bm{C}_{K-1, i, j})~~.
\end{equation}
\noindent Then, $\bm{C}^{m}_{i, j}$ is used as object priors and upsampled into the resolution of the multi-level features to scale their intensity:
\begin{equation}
\bm{\hat{X}}^{i} = \bm{X}^{i} \odot u(\bm{C}^{m})~~,
\label{eq:bd_loss}
\end{equation}
\noindent where $\odot$ and $u(\cdot)$ represent the Hadamard product and the bilinear interpolation, respectively. These scaled features are finally used as the multi-level features for the decoder $d(\cdot)$ to fuse multi-level cues and classified by the segmentation head to generate the segmentation map:
\begin{equation}
\bm{S}^{s} = s(d(\{ \bm{\hat{X}}^{1}, \bm{\hat{X}}^{2}, \bm{\hat{X}}^{3}, \bm{\hat{X}}^{4}\})~~.
\end{equation}
\noindent With the help of our OFD operation, multi-level features can be filtered based on the object priors before fusion, which can reduce invalid fusions for the feature decoding process. Thus, the segmentation map will suffer from fewer background activations on object classes, improving the performance of the segmentation branch.

\noindent \textbf{Background-aware Spatial Propagation:} In return, the segmentation branch can also provide the background score $\bm{B}$  with our proposed background-aware spatial propagation (BSP) operation to offset the background unawareness of the object classification head. As indicated in Fig.~\ref{fig:structure} (noted by yellow), our BSP assumes that the background classification map outputted by the segmentation head is a natural background score for the classification branch:
\begin{equation}
\bm{B}^{s} = \bm{S}^{s}_{0, :, :} ~~.
\end{equation}
\noindent This background score can be wrapped with the localization seed $\bm{C}$ before spatial propagation to diffuse both the object classification score and background score when generating the segmentation map of the classification branch:
\begin{equation}
\bm{S}^{c} = p(\hat{\bm{S}}^{c}) = p(\{\bm{B}^{s}; u(\mathbf{C}) \}) = p(\{\bm{S}^{s}_{0, :, :}; u(\mathbf{C}) \})~~.
\end{equation}
\noindent Compared with existing methods that generate the background score by fixed thresholds~\cite{AFA,AALR}, coarse localization maps~\cite{RRM,RRMv2}, or additional saliency maps~\cite{SEC}, our BSP can adaptively provide spatial-specific background scores with more detailed structures profited by interacting with the segmentation branch. In this way, the uncertainty caused by the unawareness of background scores can be reduced when generating the pseudo annotation $\bm{Y}^{c}$, improving the quality of segmentation maps for both branches.

\subsection{Training with Interaction Mechanisms}
We implement our interaction mechanisms with the AFA backbone structure~\cite{AFA} to make its two branches achieve mutual promotion. As shown in Fig.~\ref{fig:structure} (noted by gray), the transformer encoder is used as the feature extractor $e(\cdot)$ to generate multi-level features $\mathcal{X}=\{\bm{X}^{1}, \bm{X}^{2}, \bm{X}^{3}, \bm{X}^{4}\}$ and $m$ self-attention maps $\bm{H} \in \mathbb{R}^{m \times hw \times hw}$. These self-attentions are fed into a multilayer perceptron (MLP) layer to produce the affinity $\bm{A} \in \mathbb{R}^{hw \times hw}$ to regularize the transformer encoder with the affinity loss $\mathcal{L}_{aff}$~\cite{AFA}.

Then, to engage our interaction mechanisms, the high-level feature $\bm{X}^{4}$ is fed into the classification head to produce the localization seed $\bm{C}$ and an image-level classification score $\bm{c}$ for computing the image-level classification loss $\mathcal{L}_{cls}$. With the object prior provided by the localization seed $\bm{C}$, our OFD operation can further decode the multi-level features $\mathcal{X}$ and produce the segmentation map $\bm{S}^{s}$. Then, the background score of $\bm{S}^{s}$ also enables our BSP operation to generate another segmentation map $\bm{S}^{c}$ by propagating the localization seed with background prior. Based on the two segmentation maps, $\mathcal{L}_{c2s}$ and $\mathcal{L}_{s2c}$ are computed with our interactional supervision mechanism, and combined with $\mathcal{L}_{cls}$ and $\mathcal{L}_{aff}$ to supervise the training process:
\begin{equation}
\mathcal{L} = \mathcal{L}_{cls} + \lambda_{1} \mathcal{L}_{c2s}  + \lambda_{2} \mathcal{L}_{s2c} + \lambda_{3} \mathcal{L}_{aff} ~~,
\end{equation}
\noindent where $\lambda_{1}$, $\lambda_{2}$ and $\lambda_{3}$ are parameters to balance their affect.

\section{Experiments}
In this section, the settings of the experiments are first detailed. Then, results of our method are given for comparison with state-of-the-art methods on VOC2012 dataset and MS-COCO dataset to show the effectiveness. Finally, we conduct the ablation studies to verify each of our components. 

\begin{table}
\caption{Comparison with WSSS methods on VOC2012 dataset}
\centering	
\begin{tabular}{l|c|c|cc}
\hline
~ & Sup. & Backbone & val & test \\
\hline
\multicolumn{5}{l}{\textbf{\textit{Fully-supervised methods}}} \\
\hline
WideResNet38~\cite{RESNET38} & $\mathcal{F}$ & ResNet38 
& 77.6 & 79.7 \\
Segformer~\cite{Segformer} & $\mathcal{F}$ & MiT-B1 
& 78.7 & - \\
DeepLab~\cite{DEEPLAB} & $\mathcal{F}$ & ResNet101 
&  \textbf{80.8} & \textbf{82.5} \\
\hline
\multicolumn{5}{l}{\textbf{\textit{Multi-stages weakly supervised methods}}} \\
\hline
IRN~\cite{IRN} \footnotesize{(ICCV'19)} & $\mathcal{I}$ & ResNet101 & 63.5 & 64.8 \\
SC-CAM~\cite{SCCAM}  \footnotesize{(CVPR'20)}& $\mathcal{I}$ & ResNet101 & 61.1 & 65.7 \\
BE~\cite{BES}  \footnotesize{(ECCV'20)}& $\mathcal{I}$ & ResNet101 & 65.7 & 66.6 \\
AdvCAM~\cite{AdvCAM} \footnotesize{(CVPR'21)} & $\mathcal{I}$ & ResNet101 & 68.1 & 68.0 \\
CPN~\cite{CPN} \footnotesize{(ICCV'21)} & $\mathcal{I}$ & ResNet101 & 67.8 & 68.5 \\
ReCAM~\cite{ReCAM} \footnotesize{(CVPR'22)} & $\mathcal{I}$ & ResNet101 & 68.5 & 68.4 \\
SIPE~\cite{SIPE} \footnotesize{(CVPR'22)} & $\mathcal{I}$ & ResNet101 & 68.8 & 69.7 \\
CLIMS~\cite{CLIMS} \footnotesize{(CVPR'22)}& $\mathcal{I} + \mathcal{L}$ &  ResNet101 & 69.3 & 68.7 \\
SEAM~\cite{SEAM} \footnotesize{(CVPR'20)} &$\mathcal{I}$ &  ResNet38 & 64.5 & 65.7 \\
CDA~\cite{CDA} \footnotesize{(ICCV'21)}& $\mathcal{I}$ &  ResNet38 & 66.1 & 66.8 \\
ECS~\cite{ECS} \footnotesize{(ICCV'21)}& $\mathcal{I}$ &  ResNet38 & 68.6 & 67.6 \\
CSE~\cite{CSE} \footnotesize{(ICCV'21)}& $\mathcal{I}$ &  ResNet38 & 68.4 & 68.2 \\
SIPE~\cite{SIPE} \footnotesize{(CVPR'22)}& $\mathcal{I}$ &  ResNet38 & 68.2 & 69.5 \\
ADELE~\cite{ADELE} \footnotesize{(CVPR'22)} & $\mathcal{I}$ & ResNet38 & 69.3 & 68.8 \\
AS-BCE~\cite{ASBCE} \footnotesize{(ECCV'22)}& $\mathcal{I}$ &  ResNet38 & 68.5 & 69.7 \\
MMCST~\cite{MMCST} \footnotesize{(CVPR'23)}& $\mathcal{I} + \mathcal{L}$ &  ResNet38 & \textbf{72.2} & \textbf{72.2} \\
\hline
\multicolumn{5}{l}{\textbf{\textit{End-to-end weakly supervised methods}}} \\
\hline
EM~\cite{EM} \footnotesize{(ICCV'15)}  & $\mathcal{I}$ & VGG16 
& 38.2 & 39.6 \\
MIL~\cite{MIL} \footnotesize{(CVPR'15)}  & $\mathcal{I}$ & VGG16 
& 42.0  & 40.6 \\
CRNN~\cite{CRNN} \footnotesize{(CVPR'17)} & $\mathcal{I}$ & VGG16 
& 52.8 & 53.7 \\
1Stage~\cite{1Stage} \footnotesize{(CVPR'20)}  & $\mathcal{I}$ & ResNet38 
& 62.7 & 65.3 \\
RRM~\cite{RRM} \footnotesize{(AAAI'20)} & $\mathcal{I}$ & ResNet38 
& 62.6 & 62.9 \\
AALR~\cite{AALR} \footnotesize{(MM'21)}  & $\mathcal{I}$ & ResNet38 
& 63.9 & 64.8 \\
E2BE~\cite{E2EBE} \footnotesize{(MM'21)}  & $\mathcal{I}$ & ResNet50 
& 63.6 & 65.7 \\
RRM-v2~\cite{RRMv2} \footnotesize{(PR'22)}  & $\mathcal{I}$ & ResNet38 
& 65.4 & 65.3 \\
RRM~\cite{RRM}  \footnotesize{(AAAI'20)} & $\mathcal{I}$ & MiT-B1
 & 63.5 & - \\
AFA~\cite{AFA}  \footnotesize{(CVPR'22)} & $\mathcal{I}$ & MiT-B1 
& 63.8 & - \\
AFA + CRF~\cite{AFA} \footnotesize{(CVPR'22)} & $\mathcal{I}$ & MiT-B1 
& 66.0 & 66.3 \\
\textbf{Ours}  & $\mathcal{I}$ & MiT-B1 
& 67.2 & 67.1 \\
\textbf{Ours + CRF}  & $\mathcal{I}$ & MiT-B1 
& \textbf{68.1} & \textbf{68.1} \\
\hline
\end{tabular}
\label{tab:voc_iou}
\end{table}

\subsection{Implementation details}
ImageNet~\cite{IMAGENET} pre-trained MiT-B1~\cite{Segformer} was adopted as our transformer encoder, and the feature decoder was implemented as an All-MLP decoder~\cite{Segformer}. The classification head and the segmentation head were both implemented as  $1\times1$ convolutional operations. The pixel-adaptive refinement (PAR) strategy~\cite{AFA} was adopted as the off-the-shelf spatial propagation strategy. Random scaling with range [0.5, 2.0], random flipping, and random crop with size $512 \times 512$ are adopted to augment the input images for the training process. When generating the pseudo annotations, the image-level annotation is adopted to remove the classes that are not present in the image. Moreover, the multi-scale and flip tests with scales (0.5, 1, 1.5) were also utilized to generate the pseudo annotations. AdamW optimizer~\cite{ADAM} was used to train our model total 20,000 iterators with batch size 8 and initial learning 6e$-$5, which was then decayed with the polynomial scheduler on each training iterator. Experiments were conducted on VOC2012 dataset~\cite{VOC} and MS-COCO dataset with an Intel Core i9 CPU and four Nvidia RTX 3090 GPUs to show the effectiveness of the proposed method. 

\subsection{VOC2012 dataset}

\subsubsection{Settings} 
Following existing works~\cite{SEAM, SIPE, AFA}, we adopted the 10,582 images annotated by SBD~\cite{SBD} to train our model with only image-level annotation. The mIoU on the official training set (1,464 images) and validation set (1,449 images) were used to evaluate the localization seed and pseudo annotation. While the official validation set and test set (1,456 images) were used to evaluate the final segmentation map outputted by the segmentation branch. The confidence thresholds \{$\sigma^{c}$, $\sigma^{s}$\} were set as \{0.75, 0.5\}, and the balance terms $\{\lambda_{1}, \lambda_{2}, \lambda_{3})$ were set as \{0.7, 0.1, 0.1\}. The classification and segmentation branches were warmed up for 2,000 and 4000 iterators before adding $\mathcal{L}_{c2s}$ and $\mathcal{L}_{s2c}$ to ensure the quality of $\bm{Y}^{c}$ and $\bm{Y}^{s}$. The BSP operation was added after 4,000 iterators to replace the fixed background threshold when producing $\bm{Y}^{c}$.

\begin{table}
\centering	
\caption{Quality of localization seeds on VOC and MS-COCO train sets}
\begin{tabular}{l|c|c|cc}
\hline
~ & Backbone & Sup.  & VOC & COCO  \\
\hline
SC-CAM~\cite{SCCAM} \footnotesize{(CVPR'20)} & $\mathcal{I}$ & ResNet38
& 50.9 & - \\
SEAM~\cite{SEAM} \footnotesize{(CVPR'20)} & $\mathcal{I}$ & ResNet38
& 55.4 & 25.1 \\
CSE~\cite{CSE} \footnotesize{(ICCV'21)} & $\mathcal{I}$ & ResNet38
& 56.0 & - \\
ECS~\cite{ECS} \footnotesize{(ICCV'21)} & $\mathcal{I}$ & ResNet38
& 56.6 & - \\
CPN~\cite{CPN} \footnotesize{(ICCV'21)} & $\mathcal{I}$ & ResNet38
& 57.4 & - \\
CDA~\cite{CDA} \footnotesize{(ICCV'21)} & $\mathcal{I}$ & ResNet38
& 58.4 & - \\
AdvCAM~\cite{AdvCAM} \footnotesize{(CVPR'21)} & $\mathcal{I}$ & ResNet50
& 55.6 & 37.2 \\
SIPE~\cite{SIPE} \footnotesize{(CVPR'22)} & $\mathcal{I}$ & ResNet50
& 58.6 & - \\
AFA~\cite{AFA}  \footnotesize{(CVPR'22)} & $\mathcal{I}$ & MiT-B1
& 55.4 & - \\
MMCST~\cite{MMCST}  \footnotesize{(CVPR'23)} & $\mathcal{I} + \mathcal{L}$ & ViT-B
& 66.3 & 40.9 \\
\textbf{Ours} & $\mathcal{I}$ & MiT-B1
& \textbf{66.5} & \textbf{42.1} \\
\hline
\end{tabular}
\label{tab:voc_seed}
\end{table}

\subsubsection{Results} 
Table~\ref{tab:voc_iou} compares the quality of the segmentation results generated by our segmentation branch ($\bm{Y}^{s}$) and other state-of-the-art WSSS methods on the VOC2012 dataset. In general, profited by considering the relation between the segmentation branch and classification branch, our method achieves 68.1\% mIoU on both the VOC2012 validation and test sets. These scores are comparable with multi-stage WSSS methods and outperform all existing E2E-WSSS methods, \textit{i.e.}, 2.2\% mIoU higher than the best AFA~\cite{AFA} on average. Note that our performance is still state-of-the-art even without using the CRF post-processing to refine the segmentation map, indicating our segmentation branch can catch more comprehensive and detailed object locations. It is because the segmentation branch can feed back to the classification branch with $\mathcal{L}_{s2c}$ and provide detailed background scores $\bm{S}^{s}_{0, :, :}$, enhancing the quality of the pseudo annotation $\bm{Y}^{c}$ to promote the two branches mutually.

To further analyze the feedback effect, the quality of the localization seed generated by our classification branch and other methods are also reported in Table \ref{tab:voc_seed}. Our localization seeds achieve 66.5\% and 64.8\% mIoU,  respectively on the VOC2012 training and validation set, which are both much higher than all multi-stage and E2E-WSSS methods. As indicated in Sec.~\ref{sec:1}, the multi-stage methods cannot perceive the segmentation network, disabling them from using it to refine the seeds during the training of the classification network. Thus, they require training additional networks~\cite{IRN,AFF} to refine these low-quality seeds before generating pseudo annotations. Though E2E-WSSS methods (such as AFA) can perceive the segmentation branch when tuning the classification branch, they do not consider its feedback. So, their localization seed still has similar quality as the offline trained backbone (Baseline). Unlike them, our method considers the feedback of the segmentation branch to enhance the performance of the classification branch, making our localization seed 7.9\% mIoU higher than the best of others. 

\begin{table*}
\caption{The IoU of each category on VOC2012 validation set}
\centering
\setlength{\tabcolsep}{0.8pt}
\begin{tabular}{c|ccccccccccccccccccccccc}
~ & mIoU & bg & plane & bike & bird & boat & bottle & bus & car & cat & chair & cow & table & dog & horse & motor & man & plant & sheep & sofa & train & tv \\
\hline
SEC & 50.7 & 82.4 & 62.9 & 26.4 & 61.6 & 27.6 & 38.1 & 66.6 & 62.7 & 75.2 & 22.1 & 53.5 & 28.3 & 65.8 & 57.8 & 62.3 & 52.5 & 32.5 & 62.6 & 32.1 & 45.4 & 45.3 \\

CRNN & 52.8 & 85.8 & 65.2 & 29.4 & 63.8 & 31.2 & 37.2 & 69.6 & 64.3 & 76.2 & 21.4 & 56.3 & 29.8 & 68.2 & 60.6 & 66.2 & 55.8 & 30.8 & 66.1 & 34.9 & 48.8 & 47.1 \\

E2BE & 63.6 & 87.4 & 67.8 & 28.1 & 67.3 & 54.4 & 62.4 & 79.3 & 69.3 & 73.8 & 31.6 & 80.9 & 51.3 & 71.0 & 71.8 & 71.8 & 72.7 & 45.8 & 78.1 & \textbf{58.4} & 61.4 & 51.9 \\

1Stage & 62.7 & 88.7 & 70.4 & \textbf{35.1} & 75.7 & 51.9 & 65.8 & 71.9 & 64.2 & 81.1 & 30.8 & 73.3 & 28.1 & 81.6 & 69.1 & 62.6 & 74.8 & 48.6 & 71.0 & 40.1 & 68.5 & \textbf{64.3} \\

RRM & 62.6 & 87.9 & 75.9 & 31.7 & 78.3 & 54.6 & 62.2 & 80.5 & 73.7 & 71.2 & 30.5 & 67.4 & 40.9 & 71.8 & 66.2 & 70.3 & 72.6 & 49.0 & 70.7 & 38.4 & 62.7 & 58.4 \\

AALR & 63.9 & 88.4 & 76.3 & 33.8 & 79.9 & 34.2 & \textbf{68.2} & 75.8 & 74.8 & 82 & \textbf{31.8} & 68.7 & 47.4 & 79.1 & 68.5 & 71.4 & \textbf{80.0} & 50.3 & 76.5 & 43.0 & 55.5 & 58.5 \\

AFA  & 66.0 & 89.9 & \textbf{79.5} & 31.2 & 80.7 & \textbf{67.2} & 61.9 & 81.4 & 65.4 & 82.3 & 28.7 & 83.4 & 41.6 & 82.2 & 75.9 & 70.2 & 69.4 & \textbf{53.0} & \textbf{85.9} & 44.1 & \textbf{64.2} & 50.9 \\
\textbf{Ours} & \textbf{68.1} & \textbf{90.8} & 78.9 & 34.3 & \textbf{81.5} & 65.7 & 64.9 & \textbf{83.1} & \textbf{76.5} & \textbf{86.4} & 27.4 & \textbf{86.3} & \textbf{52.2} & \textbf{84.0} & \textbf{76.7} & \textbf{72.0} & 75.5 & 52.2 & 82.6 & 45.5 & 60.6 & 52.8 \\
\hline
\end{tabular}
\label{tab:VOC_class_val}
\end{table*}

\begin{table*}
\caption{The IoU of each category on VOC2012 test set}
\centering
\setlength{\tabcolsep}{0.8pt}
\begin{tabular}{c|ccccccccccccccccccccccc}
~ & mIoU & bg & plane & bike & bird & boat & bottle & bus & car & cat & chair & cow & table & dog & horse & motor & man & plant & sheep & sofa & train & tv \\
\hline
SEC & 51.7 & 82.4 & 62.9 & 26.4 & 61.6 & 27.6 & 38.1 & 66.6 & 62.7 & 75.2 & 22.1 & 53.5 & 28.3 & 65.8 & 57.8 & 62.3 & 52.5 & 32.5 & 62.6 & 32.1 & 45.4 & 45.3 \\
CRNN & 53.7 & 85.7 & 58.8 & 30.5 & 67.6 & 24.7 & 44.7 & 74.8 & 61.8 & 73.7 & 22.9 & 57.4 & 27.5 & 71.3 & 64.8 & 72.4 & 57.3 & 37.3 & 60.4 & 42.8 & 42.2 & 50.6 \\
E2BE & 65.7 & 88.0 & 61.8 & 27.4 & \textbf{75.3} & 46.7 & \textbf{68.0} & 76.5 & 70.2 & 79.8 & \textbf{35.1} & \textbf{81.6} & 57.6 & 80.5 & 75.6 & \textbf{81.9} & 68.2 & 55.7 & \textbf{83.4} & \textbf{60.5} & 56.5 & 50.2 \\
RRM & 62.9 & 87.8 & 77.5 & 30.8 & 71.7 & 36.0 & 64.2 & 75.3 & 70.4 & 81.7 & 29.3 & 70.4 & 52.0 & 78.6 & 73.8 & 74.4 & 72.1 & 54.2 & 75.2 & 50.6 & 42.0 & 52.5 \\
1Stage & 65.3 & 89.2 & 73.4 & \textbf{37.3} & 68.3 & 45.8 & \textbf{68.0} & 72.7 & 64.1 & 74.1 & 32.9 & 74.9 & 39.2 & \textbf{81.3} & 74.6 & 72.6 & \textbf{75.4} & \textbf{58.1} & 71.0 & 48.7 & \textbf{67.7} & \textbf{60.1} \\
\hline
Ours & \textbf{68.1} & \textbf{90.7} & \textbf{85.1} & \textbf{37.3} & 72.9 & \textbf{61.1} & 66.1 & \textbf{80.7} & \textbf{75.9} & \textbf{87.0} & 29.0 & 77.3 & \textbf{58.0} & 80.6 & \textbf{79.2} & 79.7 & 74.2 & 53.8 & 81.9 & 57.1 & 52.0 & 49.9 \\
\hline
\end{tabular}
\label{tab:VOC_class_test}
\end{table*}

Tables~\ref{tab:VOC_class_val} and Table~\ref{tab:VOC_class_test} report the comparison of different categories on the VOC2012 dataset for E2E-WSSS methods. It can be seen that our method has higher performance in nearly half of the categories, showing the effectiveness of our proposed methods. Moreover, this per-categories performance also reflects the fail cases of our method, where our method cannot well segment the class that has complex edges, such as ``bike'' and ``chair". This may be because using PAR as the spatial propagation for our BAP is still not precise as the time-consuming CRF~\cite{RRM, RRMv2} used by other methods.

\begin{figure*}
\centering
\subfigure[Visualization comparisons of images in VOC-2012 dataset]{
\includegraphics[width=0.98\textwidth]{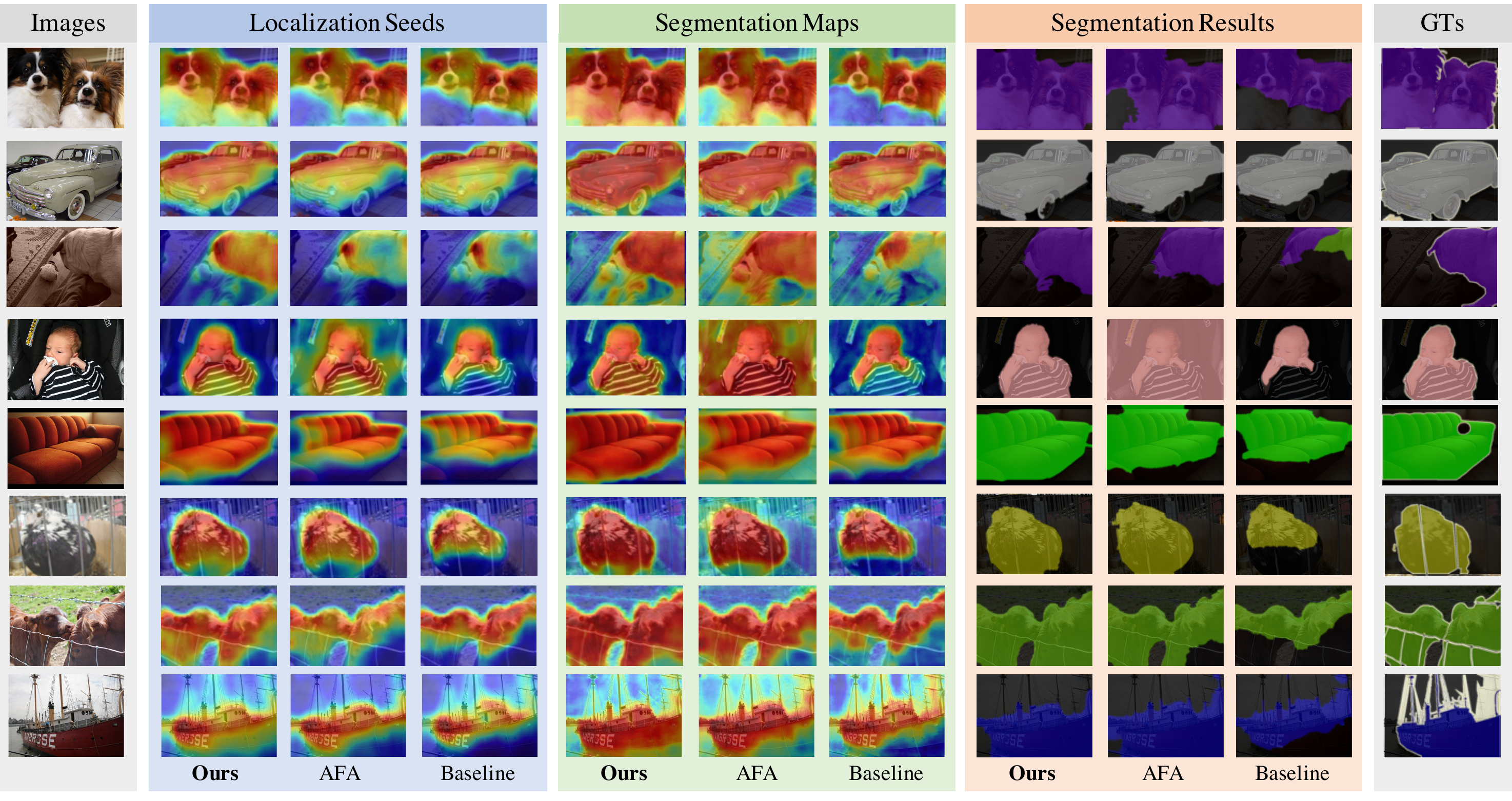}}
\subfigure[Visualization comparisons of images in MS-COCO dataset]{
\includegraphics[width=0.98\textwidth]{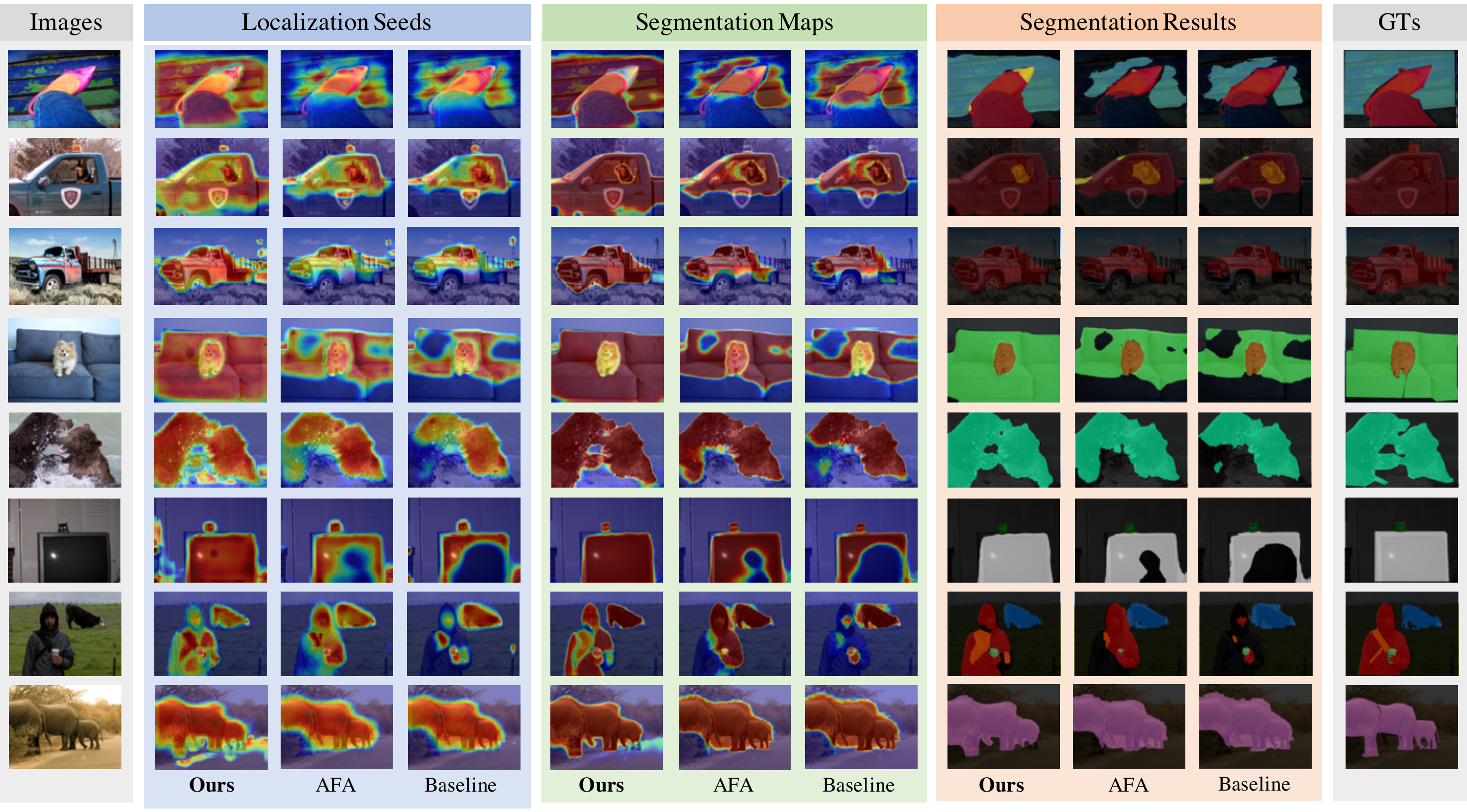}}
\caption{Visualization comparisons of localization seeds, segmentation maps, and segmentation masks generated by our mutual promotion strategy (Ours), the previous best AFA, and the baseline structure. Profited by adding interactions between two branches, our localization seed can cover more object parts with less background interference, contributing to our high-quality segmentation maps and masks.}
\label{fig:vis_map}
\end{figure*}

Fig.~\ref{fig:vis_map} also gives the visualizations comparison of different methods. It can be seen that the localization seeds of AFA only have similar activations as the baseline, indicating that the improvement of AFA is mainly attributed to its random walk score propagation with dense high-level affinity~\cite{AFA}. But, this score propagation also interferes with its performance on small objects and similar classes, such as sheep\&cow, dog\&cat, making their segmentation results contains more label noise. Compared with them, our method directly improves the quality of localization seeds to make them cover more object locations. Based on our high-quality seed, using PAR propagation with BSP to consider low-level cues is already enough to produce high-quality pseudo annotation $\bm{Y}^{c}$ to optimize the segmentation branch. Thus, our segmentation results are also better than others. 


\begin{table}
\caption{Comparison with WSSS methods on MS-COCO dataset}
\centering	
\begin{tabular}{l|c|c|c}
\hline
~ & Sup. & Backbone & val \\
\hline
\multicolumn{4}{l}{\textbf{\textit{Multi-stages weakly supervised methods}}} \\
\hline
CDA~\cite{CDA} \footnotesize{(ICCV'21)} & $\mathcal{I}$ & ResNet38
& 33.2 \\
CSE~\cite{CSE} \footnotesize{(ICCV'21)} & $\mathcal{I}$ & ResNet38
& 36.4 \\
AdvCAM~\cite{AdvCAM} \footnotesize{(CVPR'21)} & $\mathcal{I}$ & ResNet101
& 44.4 \\
AS-BCE~\cite{ASBCE} \footnotesize{(ECCV'22)} & $\mathcal{I}$ & ResNet38
& 35.2 \\
SIPE~\cite{SIPE} \footnotesize{(CVPR'22)} & $\mathcal{I}$ & ResNet38
& 40.6 \\
MMCST~\cite{MMCST} \footnotesize{(CVPR'23)} & $\mathcal{I} + \mathcal{L}$ & ResNet38
& \textbf{45.9} \\
\hline
\multicolumn{4}{l}{\textbf{\textit{End-to-end weakly supervised methods}}} \\
\hline
SLRNet~\cite{SLRNet}  \footnotesize{(IJCV'22)} & $\mathcal{I}$ & ResNet38
& 35.0 \\
AFA~\cite{AFA}  \footnotesize{(CVPR'22)} & $\mathcal{I}$ & MiT-B1 
& 38.0 \\
AFA + CRF~\cite{AFA}  \footnotesize{(CVPR'22)} & $\mathcal{I}$ & MiT-B1 
& 38.9 \\
\textbf{Ours}  & $\mathcal{I}$ & MiT-B1 
& 39.3 \\
\textbf{Ours + CRF}  & $\mathcal{I}$ & MiT-B1 
& \textbf{40.0}  \\
\hline
\end{tabular}
\label{tab:coco_iou}
\end{table}

\subsection{MS-COCO dataset}
\subsubsection{Settings} 
Except for VOC2012 dataset, experiments were also conducted on the more challenged MS-COCO dataset to show the effectiveness of our method. Following existing works~\cite{AFA}, the split of MS-COCO 2014 is used, where 82,081 images and 40,137 images of 81 classes serves as training set and validation set, respectively. $\lambda_{1}=\lambda_{2}=\lambda_3=0.1$ were used as hyper-parameters to train our method 80,000 iterators, where the first 13000 iterators are used for warmed up the classification branch. Other settings are the same as the experiments of VOC2012 dataset.

\subsubsection{Results} 
Table~\ref{tab:coco_iou} reports the performance of WSSS methods on the MS-COCO validation set. It can be seen that our methods achieve the highest segmentation performance among E2E-WSSS methods (1.1 mIoU higher than the best of others). Due to the large number of object classes and complicated context in MS-COCO images, our method still have poor segmentation performance compared with some multi-stage WSSS methods. However, as shown in Table ~\ref{tab:voc_seed}, the localization seed produce by our classification branches even have higher quality than the multi-stage methods. This indicates that our method have potential to achieve higher performance when equipped with the multi-stage training strategies~\cite{IRN, ReCAM, AdvCAM}. Except for the quantative results, the quality of localization and segmentation results on MS-COCO dataset are also given in Fig.~\ref{fig:vis_map}. The map of our method can completely catch the object locations, which is in accordance with the results of VOC2012 dataset.

\subsection{Discussion}

\subsubsection{Ablation Studies}
Ablation studies are conducted to show the effectiveness of our proposed components, including the loss functions $\mathcal{L}_{c2s}$ \& $\mathcal{L}_{s2c}$ in our interactional supervision mechanism and the BSP \& OFD operations in our branches interaction mechanism. Results are reported in Table~\ref{tab:voc_ablation}, where the performance of the segmentation maps generated by the classification and segmentation branches are represented as Branch-C ($\bm{Y}^{c}$) and Branch-S ($\bm{Y}^{s}$), respectively.

\begin{table}
\caption{Ablation Studies on VOC2012 dataset}
\centering	
\setlength{\tabcolsep}{5pt}
\begin{tabular}{cccccc|cc|cc}
\hline
\multicolumn{6}{c|}{Settings} & \multicolumn{2}{c|}{Branch-C} & \multicolumn{2}{c}{Branch-S} \\
$\mathcal{L}_{cls}$ & $\mathcal{L}_{c2s}$ & $\mathcal{L}_{s2c}$ & BSP & OFD & $\mathcal{L}_{aff}$ & train & val & val & test  \\
\hline
$\checkmark$ & ~ & ~ & ~ & ~ & ~
& 55.2 & 54.0 
& - & - \\
$\checkmark$ & $\checkmark$ & ~ & ~ & ~ & ~
& 55.7 & 54.8 
& 58.6 & 60.3 \\
$\checkmark$ & $\checkmark$ & $\checkmark$ & ~ & ~ & ~
& 62.7 & 60.9 
& 63.6 & 64.3 \\
$\checkmark$ & $\checkmark$ & $\checkmark$ & $\checkmark$ & ~ & ~
& 69.0 & 67.5 
& 65.6 & 65.4 \\
$\checkmark$ & $\checkmark$ & $\checkmark$ & $\checkmark$ &  $\checkmark$  & ~ 
& 68.1 & 66.8 
& 66.5 & 66.5 \\
$\checkmark$ & $\checkmark$ & $\checkmark$ & $\checkmark$ & $\checkmark$ & $\checkmark$ 
& 69.9 & 68.5 
& 68.1 & 68.1 \\
\hline
\end{tabular}
\label{tab:voc_ablation}
\end{table}

\begin{figure}
\centering
\includegraphics[width=0.49\textwidth]{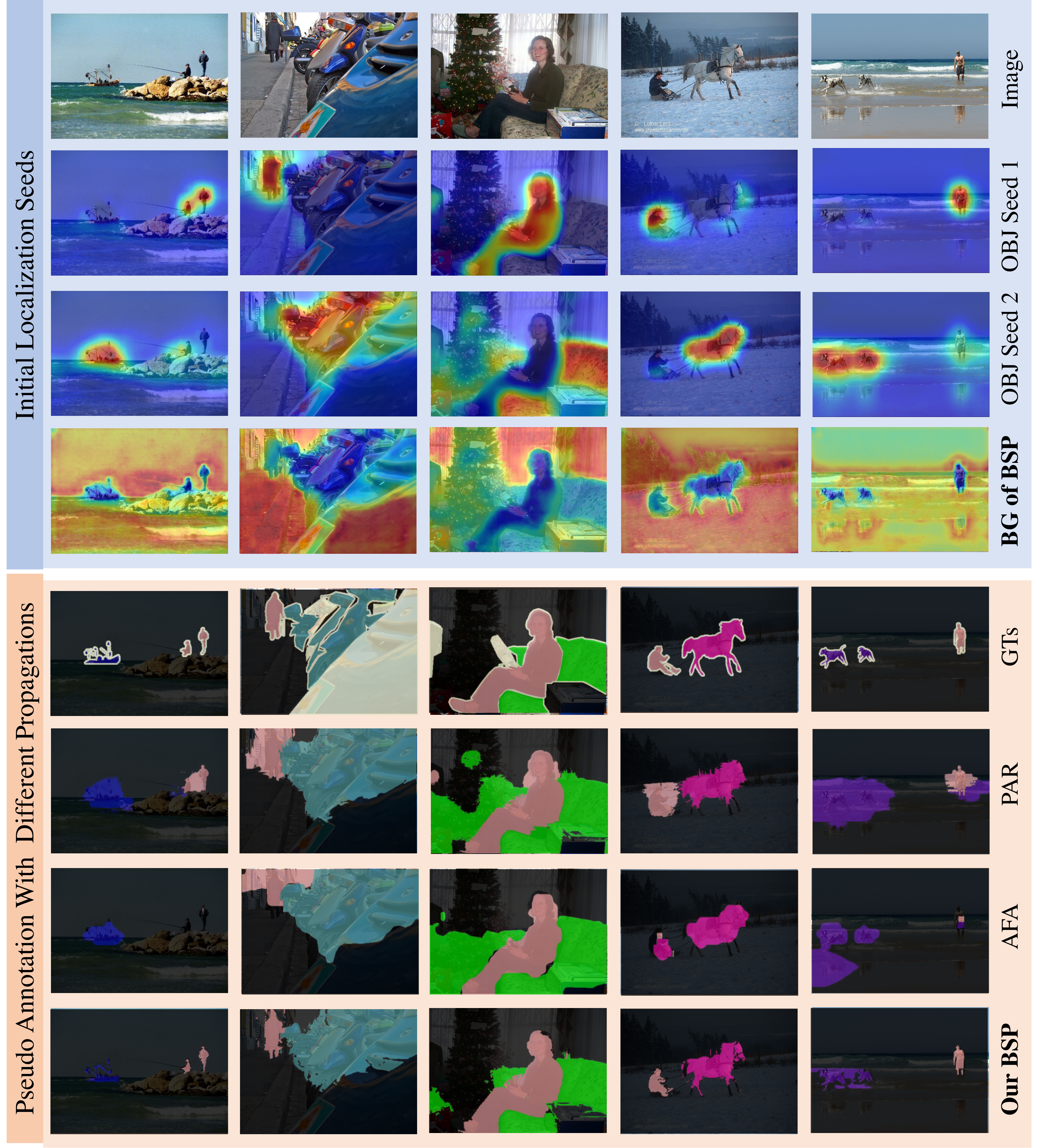}
\caption{Visualizations of the localization seeds and background priors $\bm{S}^{s}_{0, :, :}$ utilized by our BSP operation. Benefitting from our adaptive high-resolution background prior, the pseudo annotation generated by our BSP catch more detailed structure of objects.}
\label{fig:vis_bsp}
\end{figure}

As shown in Table~\ref{tab:voc_ablation}, simply using $\mathcal{L}_{c2s}$ to supervise the segmentation branch only achieves 54.8\% mIoU on the validation set, similar to the baseline offline-trained with $\mathcal{L}_{cls}$ (54.0\% mIoU). This indicates that the unidirectional supervision used by existing methods~\cite{AFA, RRM, RRMv2, AALR} limits the performance of the classification branch to provide better pseudo annotation, making the mIoU of $\bm{Y}^{s}$ also just 58.6\%. However, when interacting their supervision with $\mathcal{L}_{s2c}$ to enable their feedback, the performances of both branches are mutually promoted to a great extent, improving 6.1\% and 5.0\% respectively for $\bm{Y}^{c}$ and $\bm{Y}^{s}$ in mIoU, showing the effectiveness of our interactional supervision mechanism.

Finally, benefiting from this adaptivity, our BSP can replace PAR and propagate the localization seed better, contributing to an additional 6.6\% and 2.0\% mIoU promotion for $\bm{Y}^{c}$ and $\bm{Y}^{s}$, respectively. The qualitative comparisons between different propagation methods are also given in Fig.~\ref{fig:vis_bsp}, where our BSP operation better adapts object boundary with less background interference profited from our detailed background score. Moreover, adopting our OFD to release undesired feature fusion for the feature decoding process further improves the quality of $\bm{Y}^{s}$ into 66.5\% mIoU, which has already outperformed all existing E2E-WSSS methods. The performance will finally reach to the highest after regularizing the encoder with $\mathcal{L}_{aff}$~\cite{AFA}, achieving 68.1\% mIoU for $\bm{Y}^{s}$ on the both VOC2012 validation and test set.

\begin{figure}
\subfigure[Seed mIoU]{
\includegraphics[width=0.22\textwidth]{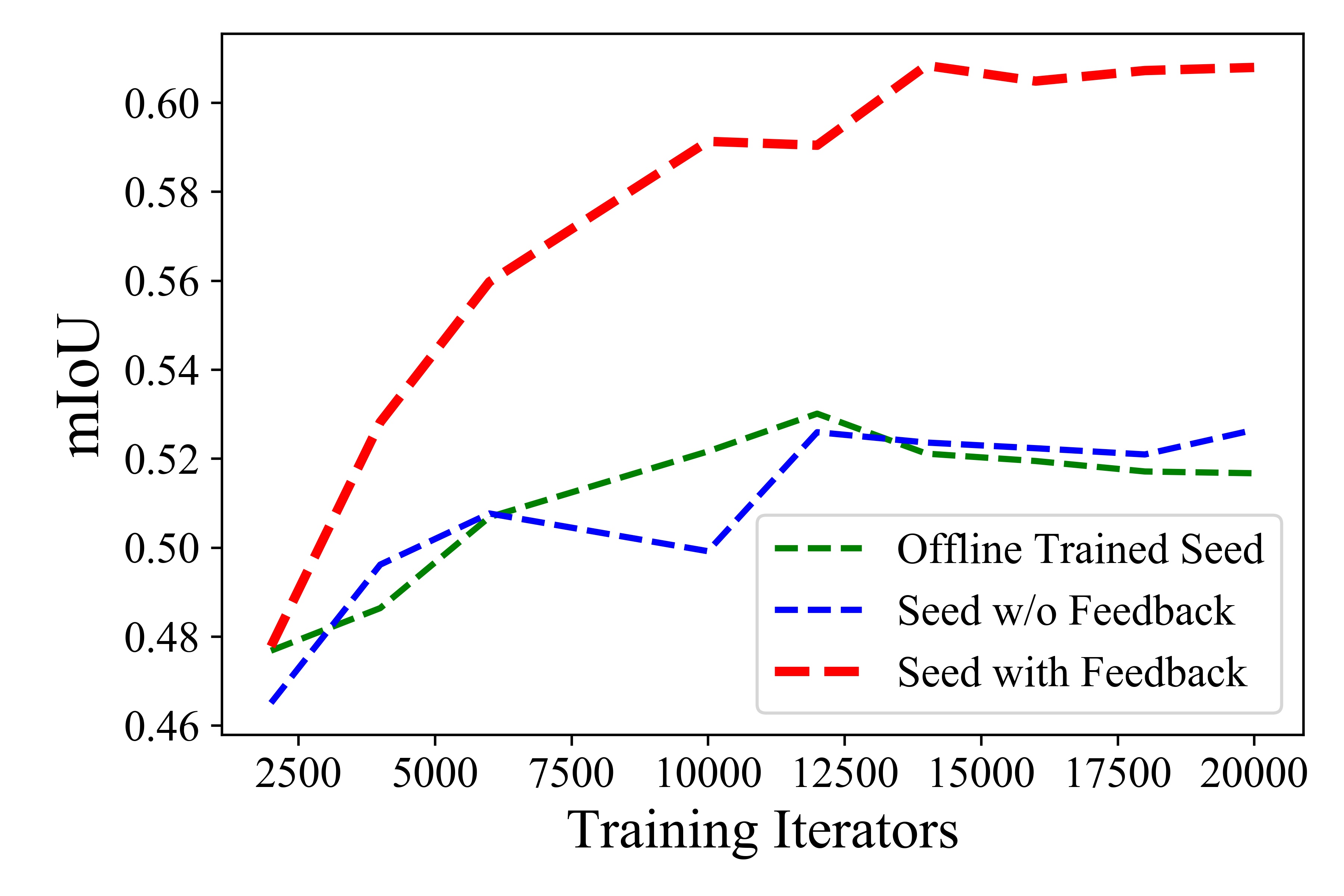}}
\subfigure[Segmentation mIoU ]{
\includegraphics[width=0.22\textwidth]{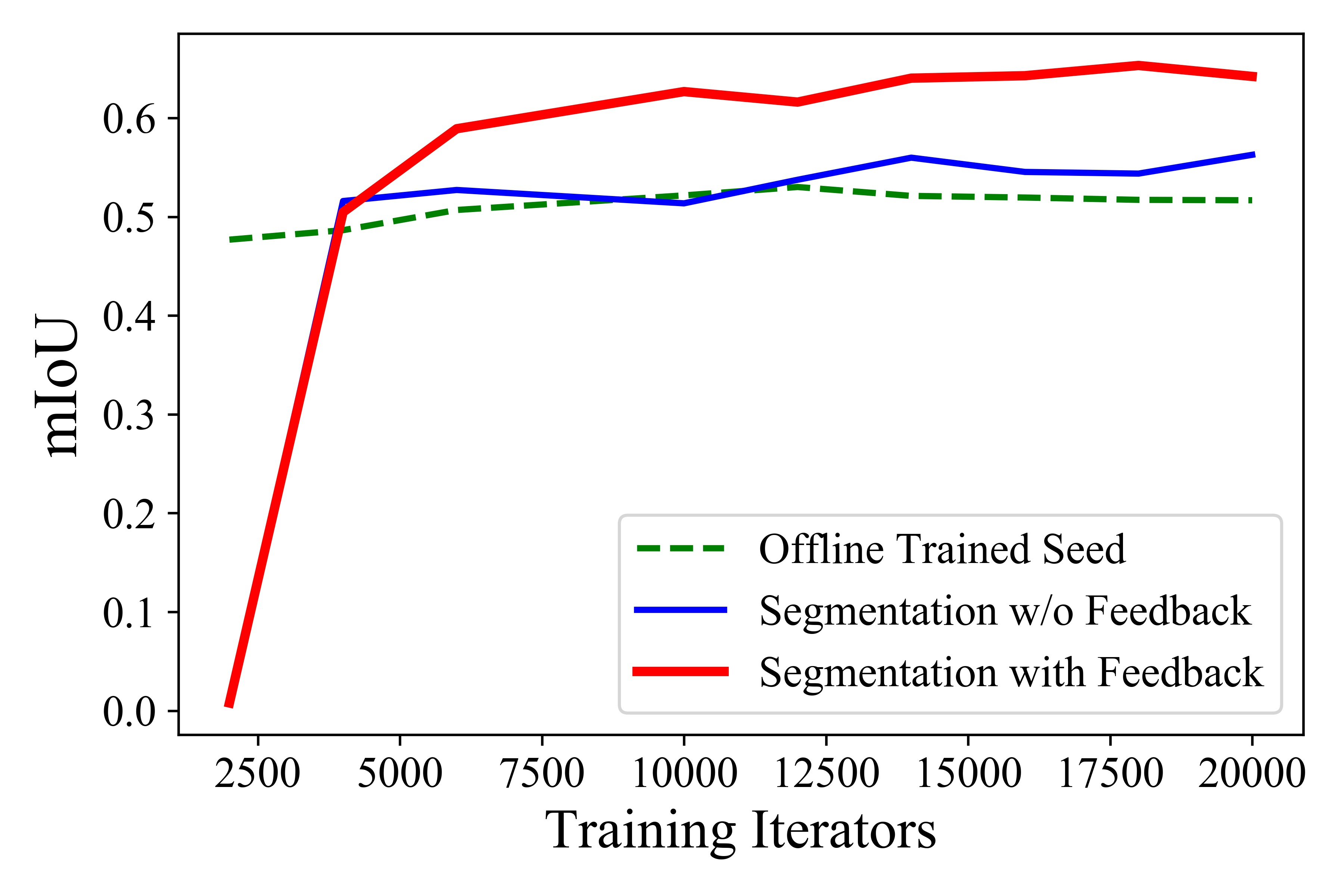}\hspace{10mm}}
\subfigure[Quality of seed of different iterators]{
\includegraphics[width=0.44\textwidth]{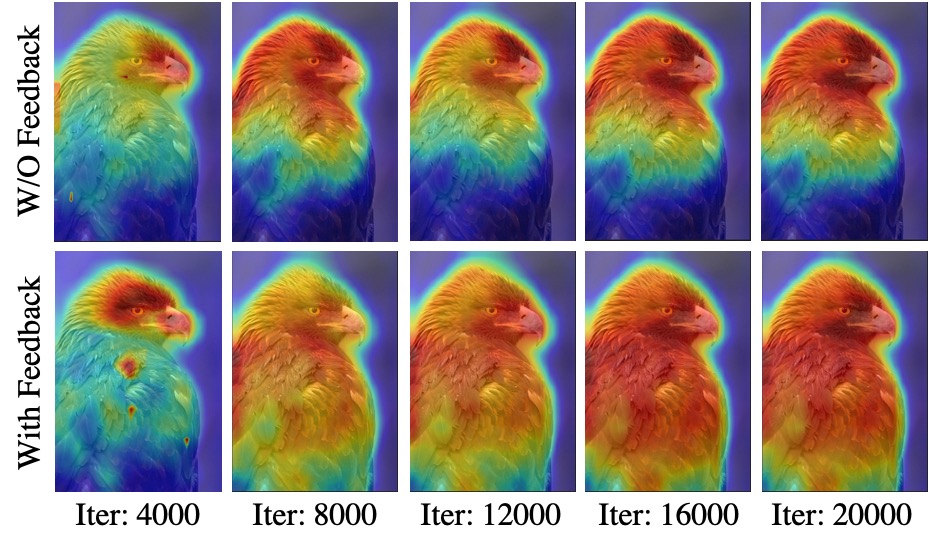}}
\caption{Results and visualizations that show the effect on each iterator when training with or w/o our mutual promotion strategy.}
\label{fig:cls_seg}
\end{figure}

\subsubsection{Confirming of Feedback Effect}
Our work assumes that the segmentation branch may perform better than the classification branch during the training process, motivating us to add interactions between these two branches for achieving mutual promotion. To confirm this fact, in Fig.~\ref{fig:cls_seg}, we give the mIoU of the localization seed (a) and segmentation map (b) respectively generated by the two branches. It can be seen that even without using our feedback loss (noted by blue), the segmentation maps outputted by the segmentation branch have higher mIoU than the localization seed after 4000 training iterators. This means that the higher performance of the segmentation branch can also be utilized to assist the classification branch in generating better localization seeds. However, these trait is not used by existing methods, making the quality of their localization seed similar to the offline-trained classification network (noted by green). Thus, after 4000 iterators, the performance of their classification branch is not obviously promoting. When using our interaction supervision mechanism (noted by red), \textit{i.e.}, adding the additional feedback loss term $\mathcal{L}_{s2c}$, the performance of the two branches can be mutually promoted during the training process. Fig.~\ref{fig:cls_seg} (c) also gives visualizations for localization seeds generated with or without using our interaction supervision mechanism. It can be seen that our interaction supervision mechanism can gradually enhance the activation of undiscriminating regions rather than overfitting the discriminative regions.

\section{Conclusions}
This work provides a novel strategy, which adds interactions between the branches of E2E-WSSS to promote them mutually. For this purpose, the interactional supervision mechanism is proposed to enable the segmentation branch give feedback to the classification branch with cross-pseudo supervision. Branches interaction mechanisms are also elaborated to add interaction operations between these two branches, making them exchange knowledges with each other. Experiments indicate our method refreshes the state-of-the-art results for E2E-WSSS on the VOC2012 dataset.

\bibliographystyle{IEEEtran}
\bibliography{egbib}

\vfill

\end{document}